\documentclass[conference]{IEEEtran}
\IEEEoverridecommandlockouts
\usepackage{cite}
\usepackage{amsmath,amssymb,amsfonts}
\usepackage{algorithmic}
\usepackage{graphicx}
\usepackage{textcomp}
\usepackage{xcolor}
\usepackage{subcaption}
\usepackage{chemformula}
\usepackage[tableposition=top]{caption}

\def\BibTeX{{\rm B\kern-.05em{\sc i\kern-.025em b}\kern-.08em
    T\kern-.1667em\lower.7ex\hbox{E}\kern-.125emX}}
\begin{document}

\title{Improving Robustness of ReRAM-based Spiking Neural Network Accelerator with Stochastic Spike-timing-dependent-plasticity\\
}

\author{\IEEEauthorblockN{Xueyuan She, Yun Long, Saibal Mukhopadhyay}
\IEEEauthorblockA{\textit{School of Electrical and Computer Engineering, Georgia Institute of Technology,  Atlanta, USA} \\
 xshe@gatech.edu, yunlong@gatech.edu, saibal.mukhopadhyay@ece.gatech.edu}

}

\maketitle

\begin{abstract}
Spike-timing-dependent-plasticity (STDP) is an unsupervised learning algorithm for spiking neural network (SNN), which promises to achieve deeper understanding of human brain and more powerful artificial intelligence. While conventional computing system fails to simulate SNN efficiently, process-in-memory (PIM) based on devices such as ReRAM can be used in designing fast and efficient STDP based SNN accelerators, as it operates in high resemblance with biological neural network. However, the real-life implementation of such design still suffers from impact of input noise and device variation. In this work, we present a novel stochastic STDP algorithm that uses spiking frequency information to dynamically adjust synaptic behavior. The algorithm is tested in pattern recognition task with noisy input and shows accuracy improvement over deterministic STDP. In addition, we show that the new algorithm can be used for designing a robust ReRAM based SNN accelerator that has strong resilience to device variation. 
\end{abstract}


\section{Introduction}

Spiking neural network (SNN) is a neuromorphic computing paradigm that mimics the behavior of animal nervous systems at the level of neurons and synapses. The development of SNN promotes deeper understanding of cognition system, and at the same time serves as a potential approach to achieving artificial neural networks (ANN) that are as efficient as the human brain, which outperforms state-of-the-art ANNs with only fraction of their power consumption~\cite{Javed2010BrainMass}. Originating from the synaptic modulation rule first studied by Hebb~\cite{Hebb1950TheTheory}, which has been the foundation of research in learning and memory ever since 1949, spike-timing-dependent-plasticity or STDP~\cite{Bi2001SynapticRevisited}, is now a widely adopted weight update algorithm in SNN. The temporal relationship between spiking events in the network as determined by STDP makes it possible for SNN to achieve learning with unlabeled input data, i.e. learning is unsupervised. In fact, SNN with STDP learning rule has been used in various computer vision related machine learning tasks~\cite{Diehl2015UnsupervisedPlasticity, Querlioz2013ImmunityNanodevices, Long2018AcceleratingApproximation} and shows accuracy results comparable with supervised neural network designs~\cite{Krizhevsky2012ImageNetNetworks}\cite{Sainath2013DeepLVCSR}.

Conventional SNN accelerators suffer from the "memory wall" of von Neumann architecture, which hinders the development of systems with higher performance as memory access speed becomes the bottleneck. One promising solution to this problem, process-in-memory, or PIM, has been actively studied in the design of novel neural network accelerators. The integration of memory and computation supports potential breakthrough of the bandwidth barrier, and can fully take advantage of the parallel operations in SNN. Resistive random access memory (ReRAM), a type of memory device that supports non-volatile modification of resistance and exhibits behavior similar to biological synapses~\cite{Jo2010NanoscaleSystems}, is an ideal candidate for such PIM SNN accelerators. ReRAM also has the advantage of high read-and-wirte speed as well as simple structure~\cite{Kawahara2013AnThroughput}. In addition, the non-volatile nature of ReRAM has the benefit of better energy efficiency as no power is need to maintain the information stored in memory.

Resistance of ReRAM is modified by voltage/current signal which changes the conducting mirco-structure between two electrodes. With controlled pulse width and amplitude of spike signal, ReRAM devices can be modified to continuous states of resistance~\cite{Lee2012MemristorApplications}, enabling analog representation of synapse conductance. Based on ReRAM, PIM design with a simple cell structure, in which only one ReRAM device is needed per synapse, can be achieved. Such design requires much less components compared to conventional CMOS implementation of SNN~\cite{Fieres2008RealizingSystem}. For instance, in Indiveri's work the circuit design for SNN with STDP based on CMOS needs about 30 transistor per synapse~\cite{Indiveri2006APlasticity}.   

Meanwhile, existing designs of ReRAM based SNN accelerators~\cite{Liu2015ACrossbar}~\cite{Saighi2015PlasticityNetworks} still face challenge caused by the stochastic nature in both the manufacturing process and operation of ReRAM. It is reported in Querlioz's work~\cite{Querlioz2013ImmunityNanodevices} that device variation of ReRAM significantly degrades learning accuracy for MNIST dateset~\cite{LeCun1998Gradient-basedRecognition}. The reason is that device variation leads to difficulties in achieving reliable STDP learning behavior as synapse conductance is stored in term of resistance of ReRAM, and each individual ReRAM device shows different resistance modulation characteristics. Beside the negative impact of device variation, noise is another factor that adversely affects learning in real life neural network applications. The event based system of SNN is sensitive to temporal and spatial distortion of spikes, and input data used in the learning process of SNN can contain noisy signal from sources such as the data collection process. In this work we propose a novel STDP learning rule and use it to implement an algorithmic approach to designing a robust ReRAM based PIM accelerator for SNN. This paper makes three key contributions:

\begin{itemize}
  \item We propose a frequency-dependent (FD) stochastic STDP algorithm based on learning rules observed in neurophysiological experiments, which can be easily integrated into most STDP based SNNs.
  \item Compared to determinstic STDP, the proposed design is able to achieve better accuracy when learning under noisy input conditions. The improvement can be observed across different noise types as well as a wide range of noise levels.
  \item The proposed algorithm shows strong resilience to the impact of device variation in ReRAM based networks, while deterministic STDP suffers from accuracy drop under the same circumstances.
\end{itemize}

In the following sections, we first discuss the fundamental theory and algorithm of SNN in section \ref{Network Models}, then present the PIM structure considered in this work in section \ref{method}. In section \ref{result} we demonstrate results from learning the MNIST dataset and compare the performance of networks based on deterministic STDP and FD stochastic STDP.

\section{Spiking Neural Network Model} \label{Network Models}
\subsection{Spiking Neuron Model}

There are different models that are developed to capture the firing pattern of real biological neurons. Long~\cite{Long2018AcceleratingApproximation} shows that with parameter tuning it is possible the achieve similar spiking frequency in different neuron models, such as Hodkin-Huxley and integrate-and-fire (LIF), when operating under a specific range of input current. This indicates that for an event based network such as the one presented in this work, which uses spikes rather than membrane potential of neurons to encode information, a mathematically less complex model can be used to achieve same level of performance as the more complex ones. To optimize speed and power consumption of our network design, we choose to use LIF model in this work. The model is described by:
\begin{equation}
dv/dt = a+bv+cI \label{eq: lif_1}\textbf{}
\end{equation}
\begin{equation}
v=v_{reset}, \text{ if } v>v_{threshold} \label{eq: lif_2}
\end{equation}
In the equation, $a$, $b$ and $c$ are decided based on the specific network settings. $I$ is the sum of current signal from all synapses that connects to neuron $m$. $I$ is evaluated by:
\begin{equation}
I_{m} = \sum_{n=0}^{N} g_{n,m}v_{pre_{n}} \label{eq: lif_3}
\end{equation}
Here $g_{n,m}$ is the conductance of the synapse connecting neuron $n$ and $m$. And $v_{pre_{n}}$ is the voltage signal resulting from spike of neuron $n$. 

\begin{figure}[!t]
\centerline{\includegraphics[width=0.48\textwidth]{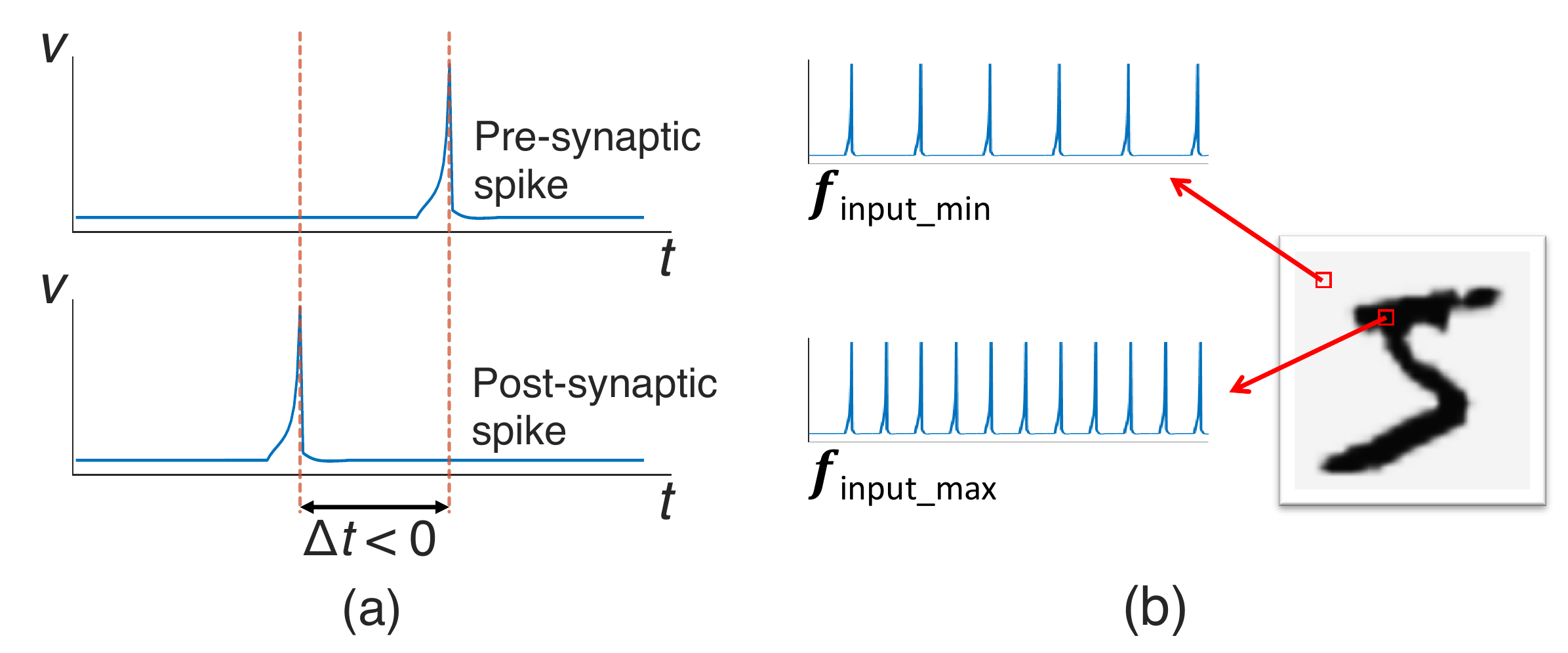}}
\caption{(a) In STDP learning rule, when pre-synaptic spikes arrive at a synapse after post-synaptic spikes, $\Delta t$ is smaller than zero. (b) Conversion of input image to spike trains: darker pixels have higher frequency 
}
\label{fig_support_1}
\end{figure}

\subsection{Synapse Model}
In SNN, two neurons connected by a synapse are called pre-synaptic neuron and post-synaptic neuron. When the pre-synaptic neuron spikes, current signal is sent through the synapse to the post-synaptic neuron. Conductance of the synapse determines how strongly two neurons are connected and can be considered the connection weight between neurons. Learning is achieved through modulating the conductance following an algorithm developed from the learning rule by Hebb~\cite{Hebb1950TheTheory} and the temporal synaptic modification rule observed in hippocampus~\cite{Bliss1973LonglastingPath}. A more detailed theoretical work by Gerstner explains learning from the perspective of spatial-temporal patterns of spikes~\cite{Gerstner1993WhyPatterns} and the algorithm is later named spike-timing-dependent-plasticity (STDP). Since then multiple experimental evidence of STDP has been observed~\cite{Markram1997RegulationEPSPs, Debanne1998Long-termCultures, Bell1997SynapticOrder, Magee1997ANeurons}, making STDP a biologically plausible algorithm that is suitable for the purpose of this work.

With STDP learning rule integrated, the network is able to extract the causality between spikes of two connected neurons from their temporal relationship. As a result, the SNN can perform conductance update without using label of the input data. More specifically, there are two operations of STDP: long-term potentiation (LTP) and long-term depression (LTD). LTP is triggered when post-synaptic neuron spikes closely after a pre-synaptic neuron spike, indicating a causal relationship between the two events, and the conductance of the synapse is increased. On the other hand, when a post-synaptic neuron spikes before pre-synaptic spike arrives or without receiving a pre-synaptic spike at all, the synapse goes through LTD which decreases its conductance. 

We choose to use STDP model presented by Querlioz~\cite{Querlioz2013ImmunityNanodevices}, as it matches experimental measurement of memresistive devices\cite{Jo2010NanoscaleSystems}\cite{Seo2011AnalogDevice} and has been tested in neural network application~\cite{Querlioz2013ImmunityNanodevices}. The model is described by the following equations:
 
\begin{equation}
\Delta G_{p} = \alpha_{p}e^{-\beta_{p}({G-G_{min}})/({G_{max}-G_{min}})} \textbf{}
\label{eq_STDP_1}
\end{equation}
\begin{equation}
\Delta G_{d} = \alpha_{d}e^{-\beta_{d}({G_{max}-G})/({G_{max}-G_{min}})} \textbf{}
\label{eq_STDP_2}
\end{equation}
$\Delta G_{p}$ is the magnitude of LTP actions, and $\Delta G_{d}$ is the magnitude of LTD actions. $\alpha_{p}$, $\alpha_{d}$, $\beta_{p}$, $\beta_{d}$, $G_{max}$ and $G_{min}$ are parameters that are tuned based on other network configurations such as input matrix size, input spiking frequency and voltage.

\subsection{Stochastic Behavior of Synapses}

In stochastic SNN, the potentiation and depression of synapses are non-deterministic, and the probability of the two actions depends on the temporal relationship between pre-synaptic and post-synaptic spikes. We consider the algorithm presented in~\cite{Srinivasan2016MagneticLearning} by Srinivasan, as probabilities are determined by:

\begin{equation}
P_{pot} = \gamma_{pot}e^{(-\Delta t/(\tau_{pot}))}
\label{eq: STOCH_STDP_1}\textbf{}
\end{equation}

\begin{equation}
P_{dep} = \gamma_{dep}e^{(\Delta t/(\tau_{dep}))}
\label{eq: STOCH_STDP_2}\textbf{}
\end{equation}

In the functions above, $\tau_{dep}$ and $\tau_{pot}$ are time constant parameters. $\Delta t$ is determined by subtracting the arrival time of the pre-synapse spike from that of the post-synapse spike ($t_{post}-t_{pre}$), as shown in Fig.~\ref{fig_support_1} (a). Probability of potentiation $P_{pot}$ is higher with smaller $\Delta t$, which indicates a stronger causal relationship. The probability of depression $P_{dep}$ is higher when $\Delta t$ is larger. $\gamma_{pot}$ and $\gamma_{dep}$ controls the peak value of probabilities.

\subsection{Proposed Stochastic STDP Model} 

\begin{figure}[!b]
\centerline{\includegraphics[width=0.48\textwidth]{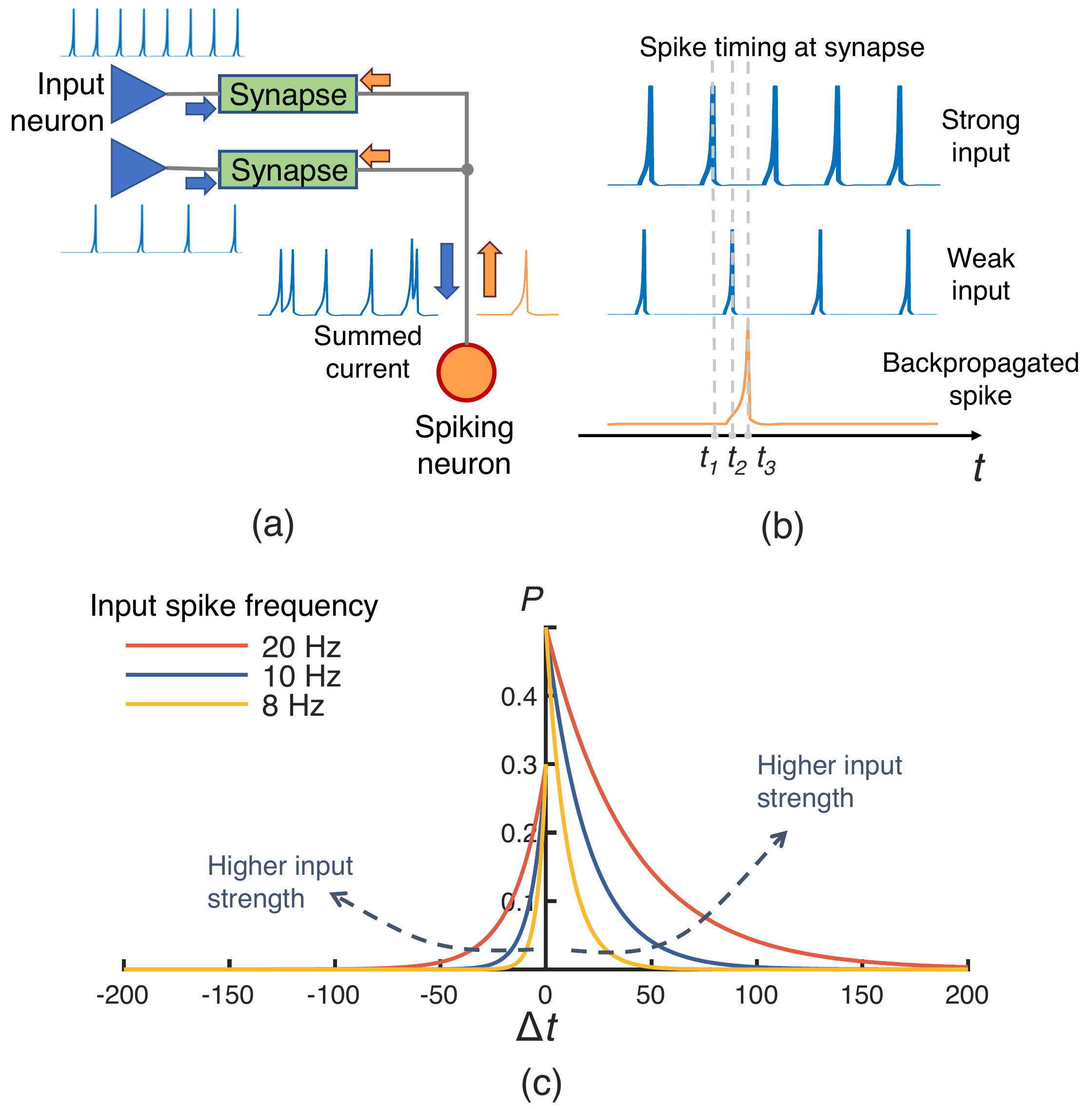}}
\caption{(a) The associative LTP behavior: convergent of two input (strong and weak) and induction of post-synaptic spike. (b) Spiking timing analysis of associative LTP. (c) Probability curves of the proposed frequency-dependent stochastic STDP with different input spike frequency.
}
\label{fig_associative}
\end{figure}

Srinivasan's algorithm captures the exponential dependence on timing of synaptic behavior as observed in biological experiments~\cite{Bi2001SynapticRevisited}. However it falls short to resolve the associative potentiation issue in STDP, which is directly related to the theoretical construct of the Hebbian synapse~\cite{Hebb1950TheTheory}. Associativity in biological synaptic behavior is first reported by Levy~\cite{Levy1979SynapsesFormation} and is proved to be an important role in the forming of classical conditioning of nervous systems~\cite{Carew1981ClassicalCalifornica}~\cite{Hawkins1983AFacilitation}. As shown in Fig.~\ref{fig_associative} (a), associativity is a temporal specificity such that when a strong (in case of our SNN model, more frequent) input and a weak (less frequent) input into one neuron induce a post-synaptic spike, a following conductance modulation process is triggered. For timing of spikes received by the synapse $t_1$, $t_2$ and $t_3$, as shown in Fig.~\ref{fig_associative} (b), $\Delta t = t_3-t_2$ for the weak input is smaller than $\Delta t' = t_3-t_1$ for the strong input. The occurrence of the post-synaptic spike in this case has more correlation to the strong input, whereas the coincidental timing of spikes from the weak input presents an "illusive" causal relationship. Experimental result from Levy~\cite{Levy1983TemporalHippocampus} shows that in hippocampus, if the weak input spike arrives before (by as much as 20 ms) or at the same time with the strong input spike, LTP of the synapse transmitting the weak input spike is induced. If the weak input spike arrives after the strong input spike, LTD is induced. In this way the nervous system detects the "illusive" event and react properly. 

In the context of STDP based SNNs, associativity can cause erroneous conductance modulation if unaccounted for. Therefore, we propose a frequency-dependent (FD) stochastic STDP that dynamically adjust the width of LTP/LTD window based on input signal frequency. The algorithm is now described by:

\begin{equation}
P_{pot} = \gamma_{pot}e^{(-\Delta t/(\tau_{pot}(1+\Phi_{pot})))}
\label{eq: STOCH_STDP_new_1}\textbf{}
\end{equation}

\begin{equation}
P_{dep} = \gamma_{dep}e^{(\Delta t/(\tau_{dep}(1+\Phi_{dep})))}
\label{eq: STOCH_STDP_new_2}\textbf{}
\end{equation}

\begin{equation}
\Phi_{dep} = \phi_{dep}\frac{f-f_{min}}{f_{max}-f_{min}}
\label{eq: STOCH_STDP_new_4}\textbf{}
\end{equation}

\begin{equation}
\Phi_{pot} = \phi_{pot}\frac{f-f_{min}}{f_{max}-f_{min}}
\label{eq: STOCH_STDP_new_5}\textbf{}
\end{equation}

Shown in Fig.~\ref{fig_associative} (c) are probability curves of this algorithm. When input spike originates from a weak input, the probability declines faster than that from a strong input. As a result, spike arriving time for weak input needs to be much closer to the post-synaptic spike to have the same probability of inducing LTP, i.e. the window for LTP is narrower for weak input. The same rule applies to LTD behavior.
\begin{figure}[!t]
\centerline{\includegraphics[width=0.39\textwidth]{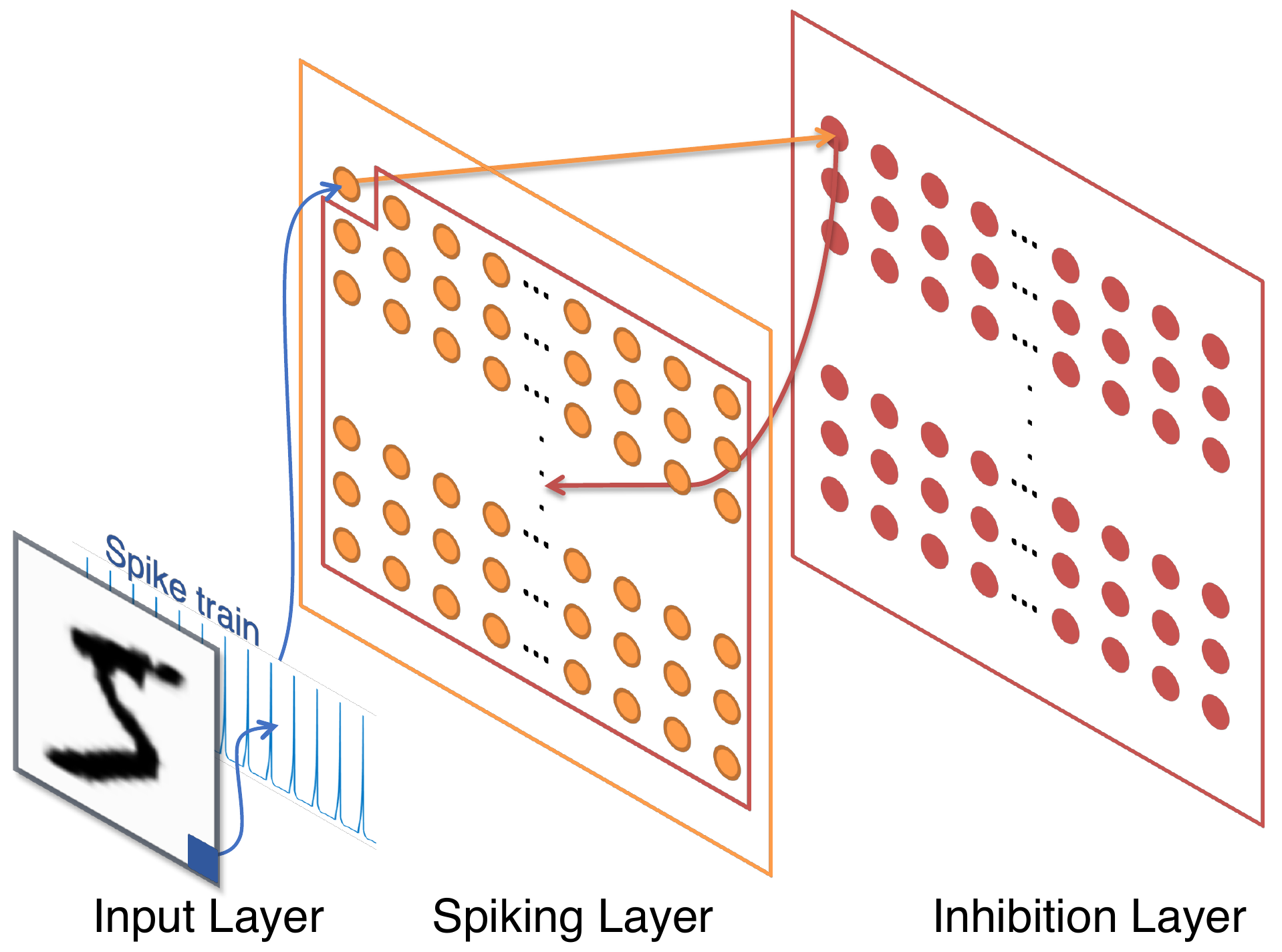}}
\caption{Network architecture of the SNN implemented in this work.
}
\label{fig_archi}
\end{figure}
\subsection{Network Architecture} 

In this work we use an SNN architecture as shown in Fig.~\ref{fig_archi}. This architecture is designed for pattern recognition tasks and consists of three layers. First is the input layer. Each neuron in this layer corresponds to one pixel in the input image. During learning process, the 8-bit pixel intensity from the input data is converted into spiking frequency over a range from $f_{min}$ to $f_{max}$, as shown in Fig.~\ref{fig_support_1} (b), and the relationship is direct proportional. Each input unit keeps track of the corresponding spiking frequency and during learning period $t_{learn}$ of one image, constantly sends excitatory spike signals to the next layer at such frequency. The input layer connects to the spiking neuron layer (second layer) in an all-to-all fashion (fully connected). The inhibition layer (third layer) has the same dimension as the second layer. For a neuron in the third layer at location $<i, j>$, there is only one input connection to it which is from the neuron at the same location, i.e. $<i, j>$, in the second layer, and its output connects to all second layer neurons except for $<i, j>$. 

As a result, when one neuron in the second layer spikes, it sends excitatory signal to the one corresponding neuron in the third layer. The inhibitory neuron has low threshold that it activates immediately after receiving one excitatory signal, and then sends inhibitory signal to all other neurons in the second layer for a period of time $t_{inh}$. Membrane potential of neurons that receive the inhibitory signal is decreased by a value of $v_{inh}$, and can not spike during $t_{inh}$. With the inhibition layer implemented, the network achieves a winner-take-all principle throughout the spiking neuron layer, preventing multiple neurons from learning the same pattern. STDP is applied to connections between the first and second layer. The conductance matrix of synapses connected to a single spiking neuron forms a learned template that contains features of one pattern.

\section {Model of Hardware Architecture} \label{method}

\subsection{ReRAM based SNN Accelerator Design} 
\begin{figure}[!b]
\centerline{\includegraphics[width=0.5\textwidth]{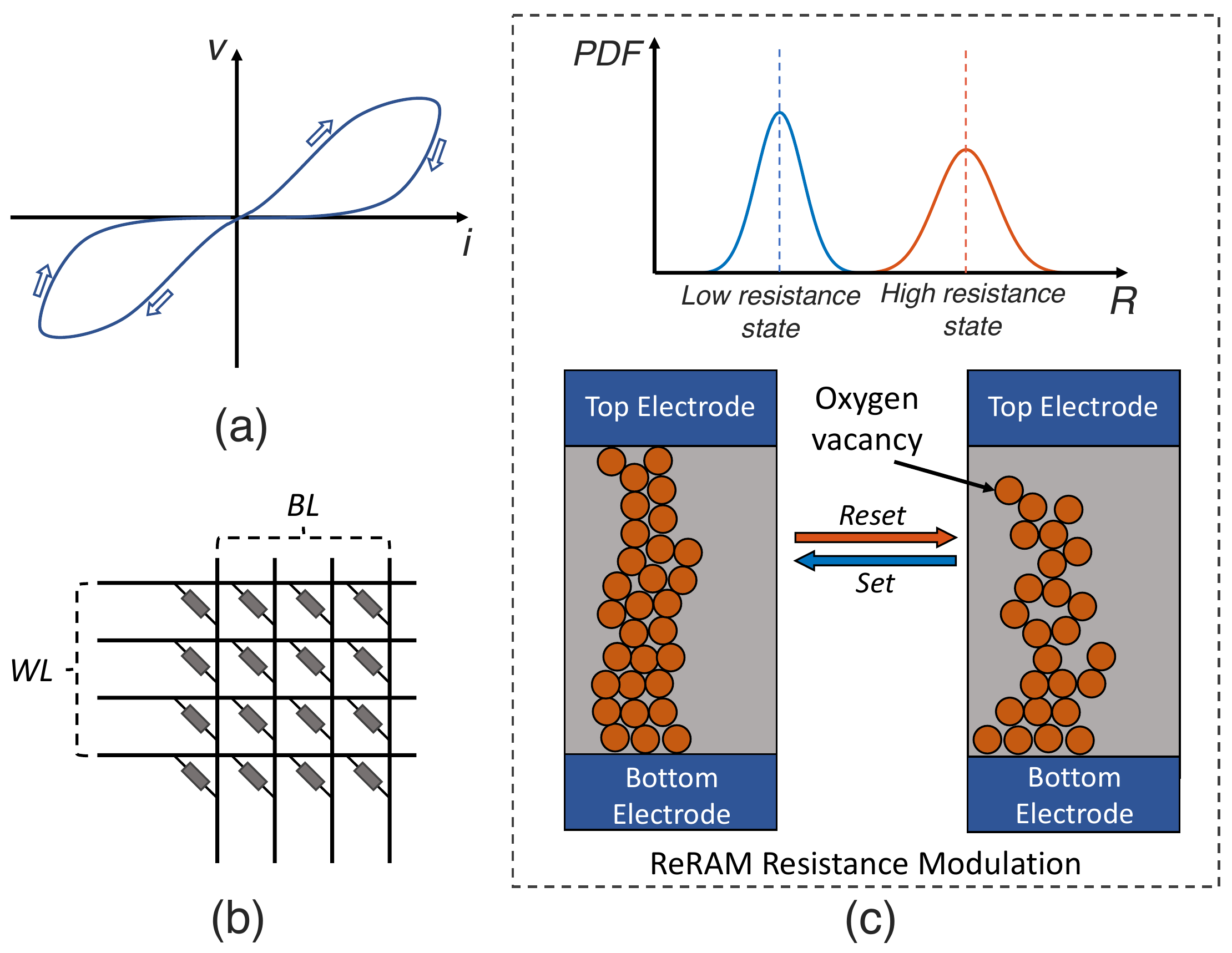}}
\caption{(a) ReRAM voltage-current characteristic with the "pinched at zero" curve. (b) ReRAM crossbar array. (c) Probability density function (PDF) of ReRAM resistance distribution at low resistance state and high resistance state (top); the change of conductive filaments geometry during resistance modulation of ReRAM (bottom): as gaps forms between oxygen vacancies, resistance of ReRAM increases.
}
\label{fig_reram_2}
\end{figure}

ReRAM is a two terminal device where the resistive switching layer (e.g. \ch{HfOx}, \ch{NiO}, \ch{TiO2}, \ch{Al2O3}, or their combinations) sandwitched between the top and bottom electrodes as shown in Fig.~\ref{fig_reram_2} (c). Device resistance can be modulated by applying set/reset voltage with different directions. In general, ReRAM can be defined as devices that exhibit a voltage-current characteristic~\cite{0268-1242-29-10-104001} as shown in Fig.~\ref{fig_reram_2} (a). Pinched at 0 Volt, the curve shows resistance variation of the ReRAM device under different injection current. Specifically, the set process happens when a positive voltage is applied on the top electrode which causes the oxygen ion migration, leaving oxygen vacancy to form a conductive filament (CF). On the other hand, device is reset by applying a reversed voltage which causes the recombination of oxygen ion and vacancy. The formation and rupture of conductive filament determines the device resistance, and thus, different data storage status.

The non-volatile memory and synapse like resistance modulation process of ReRAM make it an ideal candidate for STDP based SNN circuit. We consider the crossbar structure that has been explored in several ReRAM based SNN hardware implementations~\cite{Querlioz2013ImmunityNanodevices}\cite{Liu2015ACrossbar}\cite{Saighi2015PlasticityNetworks} to be a good paradigm of designing SNN accelerators. As shown in Fig.~\ref{fig_reram_2} (b), in this crossbar array wordlines (WL) and bitlines (BL) are connected by ReRAM at intersections.

\begin{figure}[!t]
\centerline{\includegraphics[width=0.48\textwidth]{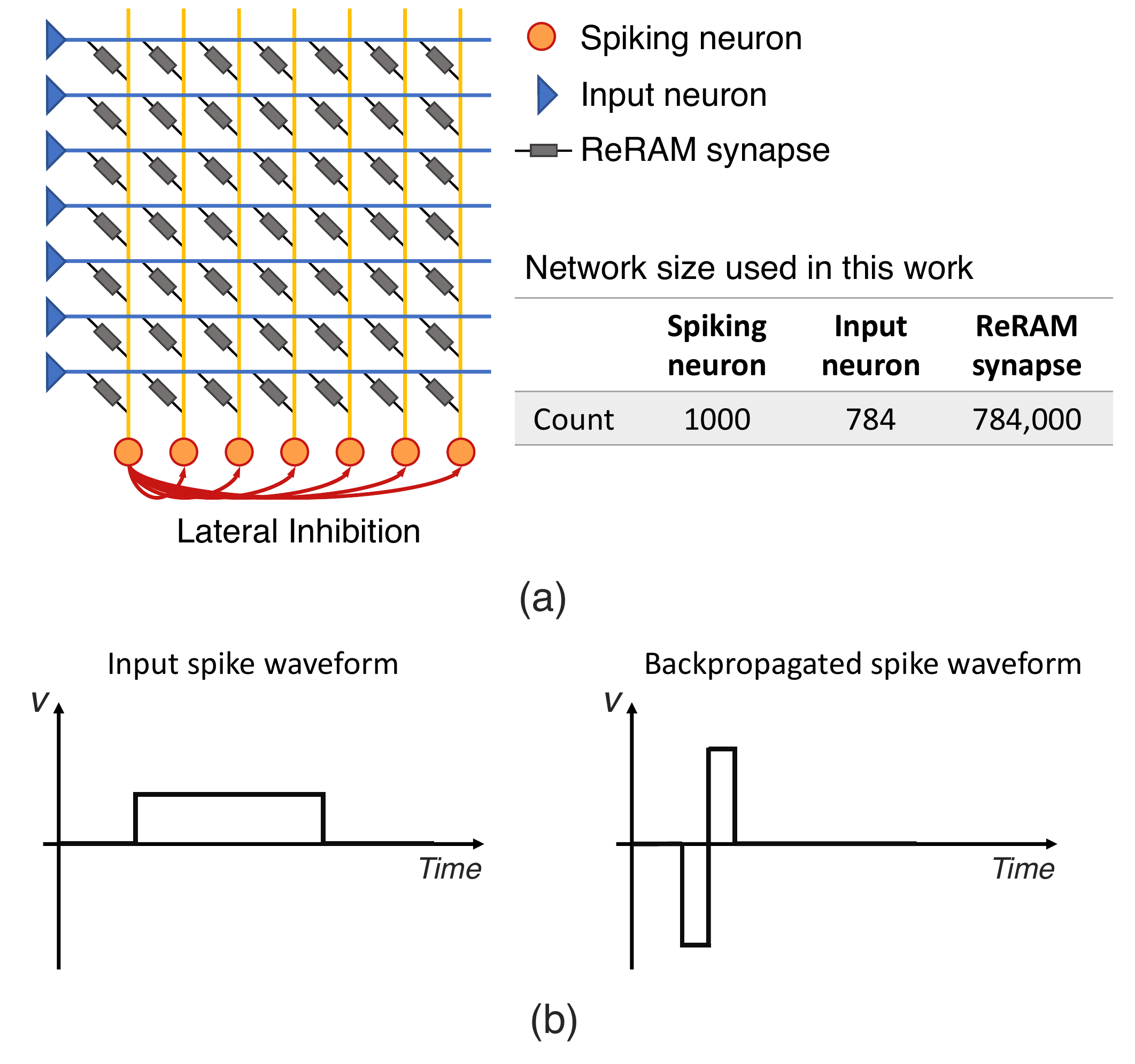}}
\caption{(a) Crossbar array design of SNN circuit with ReRAM as synapses. (b) Waveform of input spike (signal from input neuron) and backpropagated spike (signal from spiking neuron).
}
\label{fig_reram_1}
\end{figure}

\subsection{Mapping SNN into ReRAM Accelerator}
Based on the SNN architecture shown in Fig.~\ref{fig_archi}, an ReRAM crossbar PIM accelerator can be constructed accordingly for pattern recognition tasks. In our simulation that uses MNIST dataset, which has an image dimension of 28X28, a total of 784 input neurons are connected to WL. While with more spiking neurons the network is able to learn more distinct templates and therefore achieve better classification accuracy, in this work we consider a balanced option between area and performance, and use a second layer with 1000 spiking neuron. It is possible to use ReRAM as binary device by separating the high resistance state and low resistance state. However it would require more devices for each synapse to achieve enough bit width. Here we choose to utilize the continuous resistance states of ReRAM so that only one analog device is need per synapse. With ReRAM placed at each intersection of the crossbar structure, an all-to-all connection is achieved between input neuron and spiking neuron. The result network is shown in Fig.~\ref{fig_reram_1} (a). During operation, spike trains (pre-synaptic spikes) from input neurons are sent through the horizontal lines. Signal from each individual synapse on one vertical line is collected and the sum sent to the receiving neuron. If membrane potential of the receiving neuron exceeds its spiking threshold, a post-synaptic spike (backpropagated spike) is sent backwards into the vertical line. The waveform of input spikes and backpropagated spikes used in this design, as demonstrated in Fig.~\ref{fig_reram_1} (b), has been shown in\cite{Querlioz2013ImmunityNanodevices} to create resistance modulation behavior on ReRAM as described by \eqref{eq_STDP_1} and \eqref{eq_STDP_2}.

\subsection{Device Variation in ReRAM}
Recent study of ReRAM shows that device variations is one of its intrinsic properties~\cite{inproceedings}. Such variations is a result of the unique resistance modulation operation of ReRAM. In different resistance states of ReRAM distinct conduction mechanisms dominate. At low resistance state, the conduction is similar to metallic conductors as CF constitutes of dense oxygen vacancies. On the other hand, at high resistance state, gaps form between vacancies and electrons move in hopping motion as described by\cite{Mott1979ElectronicMaterials}. In case of ultra-high resistance state, connections between CF and electrode are broken and electrons need to overcome large tunneling barrier during conduction. 

The resistance modulation process thus changes the fundamental conduction mechanism of an ReRAM device across resistance states, and its randomness originates from a few different sources. First, in order to change device resistance, current impulses are injected through the ReRAM in a process called electric formation~\cite{Wei2017AnalyticEquation}. During electric formation oxygen vacancies generation and redistribution happens randomly in the crystal structure, as a result of random walk~\cite{Grimmett1993RandomModel}. Those vacancies constitute CFs that are in stochastic geometry and have different resistance. In addition to that, electric formation also creates different gaps between vacancies that affect electron hopping conductivity. Similarly, it leads to varying tunneling distance between tip of CF and device terminals~\cite{Yu2011OnCharacterization} in ultra-high reistance state. Therefore, a fixed current impulse can produce different resistance modulation among ReRAM devices and even across operation cycles on the same device. Those variations lead to a statistical distribution of device resistance rather than a specific value. Lin and Li finds in their work that ReRAM device variation can be characterized by a normal or log-normal distribution~\cite{Lin:2018:DSF:3240765.3240800}\cite{Li2015Variation-AwareModel}. 

Prior works have demonstrated that the intrinsic device variation can be a major concern in terms of computing accuracy~\cite{Long2019DesignReRAM}. In this paper, we test the proposed FD stochastic STDP algorithm for enhancing the system robustness for device variation. When simulating the SNN, we model such variation by applying parametrized Gaussian noise to the conductance value after each resistance modification process. Specifically, the distribution of an expected $G$ is:
\begin{equation}
X_{G} \sim N(G, \sigma_{dv}^2 = (\gamma_{dv}G)^2)
\end{equation}
The distribution is centered at the expected value, with standard deviation proportional to the expected $G$. And $\gamma_{dv}$ is a parameter used to control variation level.

\begin{figure}[!b]
\centerline{\includegraphics[width=0.47\textwidth]{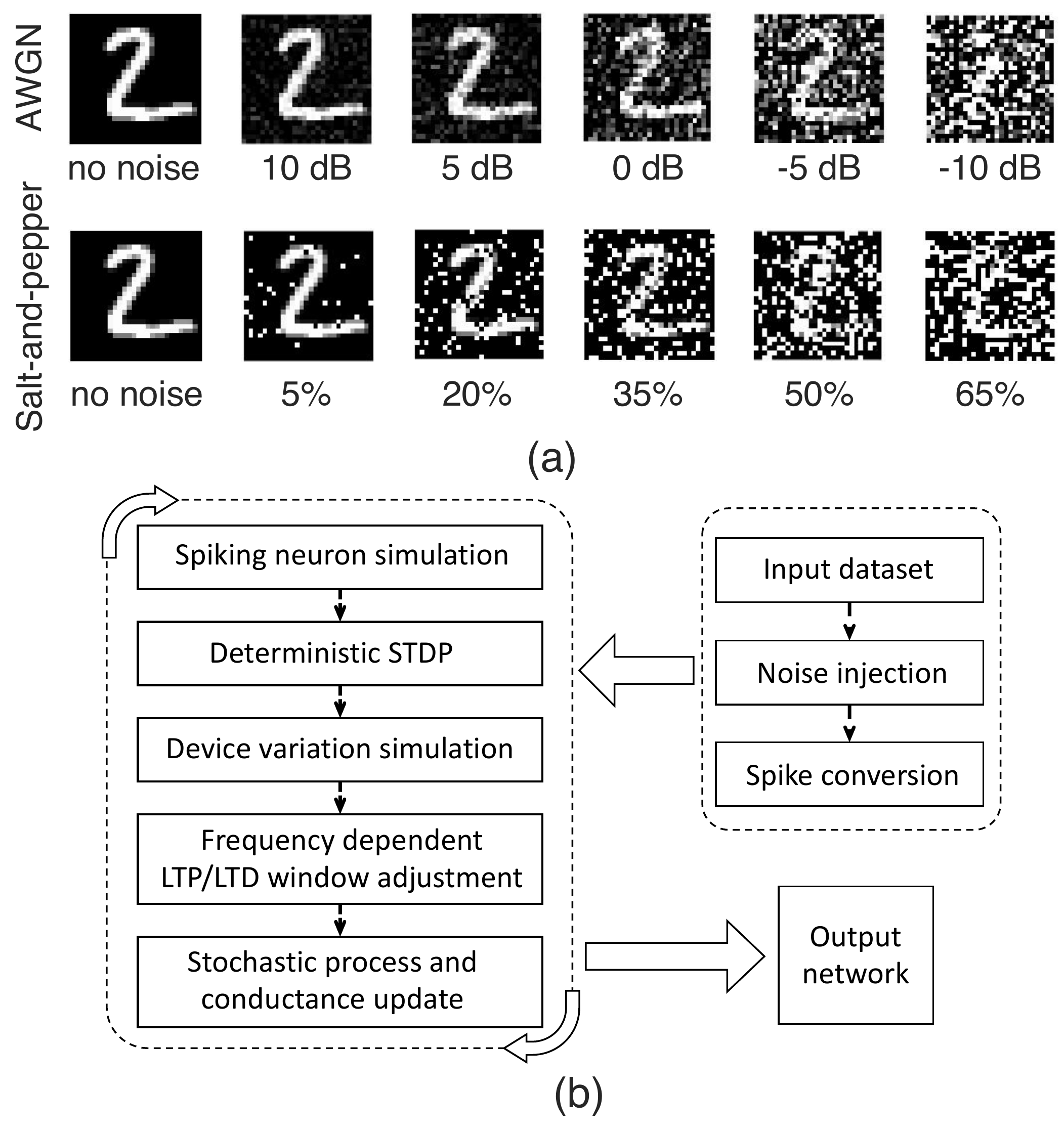}}
\caption{(a) Examples of different level of AWGN and salt-and-pepper noises used in this work. (b) Program flow of this SNN simulator.
}
\label{fig_support_2}
\end{figure}

\section{Results} \label{result}
\begin{figure*}[]
\centerline{\includegraphics[width=1\textwidth]{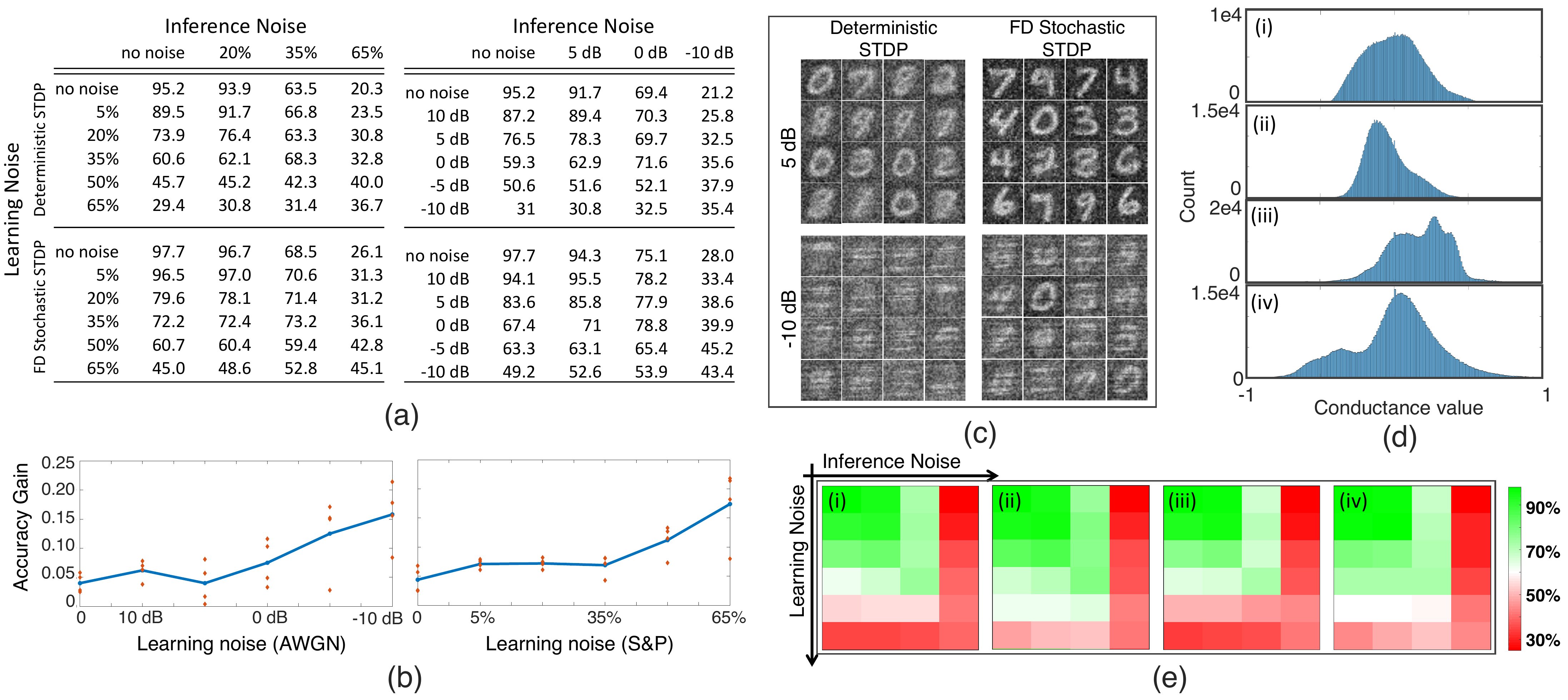}}
\caption{(a) Accuracy of learning input with AWGN noise (left) and with salt-and-pepper noise (right). (b) Accuracy gain of FD stochastic STDP over deterministic STDP for AWGN noise (left) and salt-and-pepper noise (right); red dots represents results from different inference noise levels and blue line tracks average accuracy gain over tests on all inference noise levels. (c) Visualization of patterns learned from 5 dB and -10 dB input dataset by two network configurations. (d) Distribution of synapse conductance for (i) deterministic STDP learning 5 dB noise input, (ii) FD stochastic STDP learning 5 dB noise input, (iii) deterministic STDP learning -10 dB noise of input, and (iv) FD stochastic STDP learning -10 dB noise input. (e) Heat maps of classification accuracy of (i) deterministic STDP learning input with AWGN noise, (ii) FD stochastic STDP learning input with AWGN noise, (iii) deterministic STDP learning input with salt-and-pepper noise and (iv) FD stochastic STDP learning input with salt-and-pepper noise.}
\label{fig_noise_1}
\end{figure*}

\subsection{Simulation Settings}
In order to test the SNN performance under noisy input condition, noise is added to the original image data prior to the learning process. As shown in Fig.~\ref{fig_support_2} (a), two types of noise: Additive white Gaussian noise (AWGN) and salt-and-peper, with different noise levels are tested. AWGN noise is measured in Signal-to-noise ratio (SNR). For input image to spike train conversion, $f_{min}$ is 5 Hz and $f_{max}$ is 22 Hz. For the LIF model $V\_th$ is -60.2, $V\_reset$ is -74.7, $a$ is -6.77, $b$ is -0.0989 and $c$ is 0.314. For the base STDP algorithm, $\alpha_{P}$ is 0.01, $\beta_{P}$ is 3, $\alpha_{D}$ is 0.005, $\beta_{D}$ is 3, $G_{max}$ is 1.0 and $G_{min}$ is 0. For FD stochastic STDP, $\gamma_{pot}$ is 0.3, $\tau_{pot}$ is 80, $\gamma_{dep}$ is 0.2, $\tau_{dep}$ is 5, $\phi_{dep}$ is 0.3 and $\phi_{pot}$ is 0.1. The MNIST dataset contains 60,000 training images and 10,000 test images. For each learning process all 60,000 training images are shown to the network. The first 1000 images in the test set are used to label all learned patterns and the remaining 9,000 are used to test classification accuracy. In this work we use a GPU accelerated spiking neural network simulator called ParallelSpikeSim\cite{She2019FastNetwork} running on a desktop machine with Intel Core i7-7700K and NVIDIA GTX 1080 Ti GPU. The program flow is shown in Fig.~\ref{fig_support_2} (b). 

\subsection{Learning with Noisy Input Data}
For the two types of noise tested, FD stochastic STDP based network shows better performance in almost all test conditions as shown in Fig.~\ref{fig_noise_1} (a). Accuracy gain at each learning noise level is shown in Fig.~\ref{fig_noise_1} (b) as discrete points and average accuracy gain over all inference noise conditions is shown as a line. It can be observed that accuracy improvement over network based on deterministic STDP is more prominent when learning input is more noisy, e.g. with non-noisy input learning the average of all test cases is 2.9\% higher for FD stochastic STDP, while at 0 dB noise the improvement rises to 8.1\%. A similar trend exists for salt-and-pepper noise, as average accuracy improvement increases from 2.7\% at non-noisy input to 8.6\% at 35\% noise. It is also worth noting that networks which learn noisy input dataset show better performance at classifying test images with noise than those with no noise, and maximum accuracy result of a specific inference noise level comes from networks that receives similarly noisy input for learning. This is illustrated in the heatmaps from Fig.~\ref{fig_noise_1} (e). Each heatmap shows accuracy in one simulation setting (e.g. FD Stochastic STDP learning input with AWGN noise). The x-axis represents inference noise and y-axis represents test noise; higher accuracy is displayed in green and lower in red. The diagonal positions of each heatmap have higher accuracy than neighbor positions.

Visualization of conductance matrix after learning input with AWGN noise is shown in Fig.~\ref{fig_noise_1} (c). At 5 dB input noise both deterministic and FD stochastic STDP are able to learn patterns with distinct features of different digits. Fig.~\ref{fig_noise_1} (d) shows the distribution of conductance of all 786,000 synapses in the network, with deterministic STDP and FD stochastic STDP both showing distribution resembling normal distribution. Indeed, when looking at the extreme case where SNR decreases to -10 dB, learning abilities of two network configurations are negatively affected. However, with FD stochastic STDP, the network can extract more features from the noisy input, and achieve around 20\% higher classification accuracy. This performance difference can also be reflected in the distribution of conductance in the network as shown in Fig.~\ref{fig_noise_1} (d). The distribution of deterministic STDP, which has a wide and flat region, is less ideal than that of FD stochastic STDP which is closer to normal distribution.

\subsection{Network with ReRAM Device Variation}
\begin{figure}[!t]
\centerline{\includegraphics[width=0.46\textwidth]{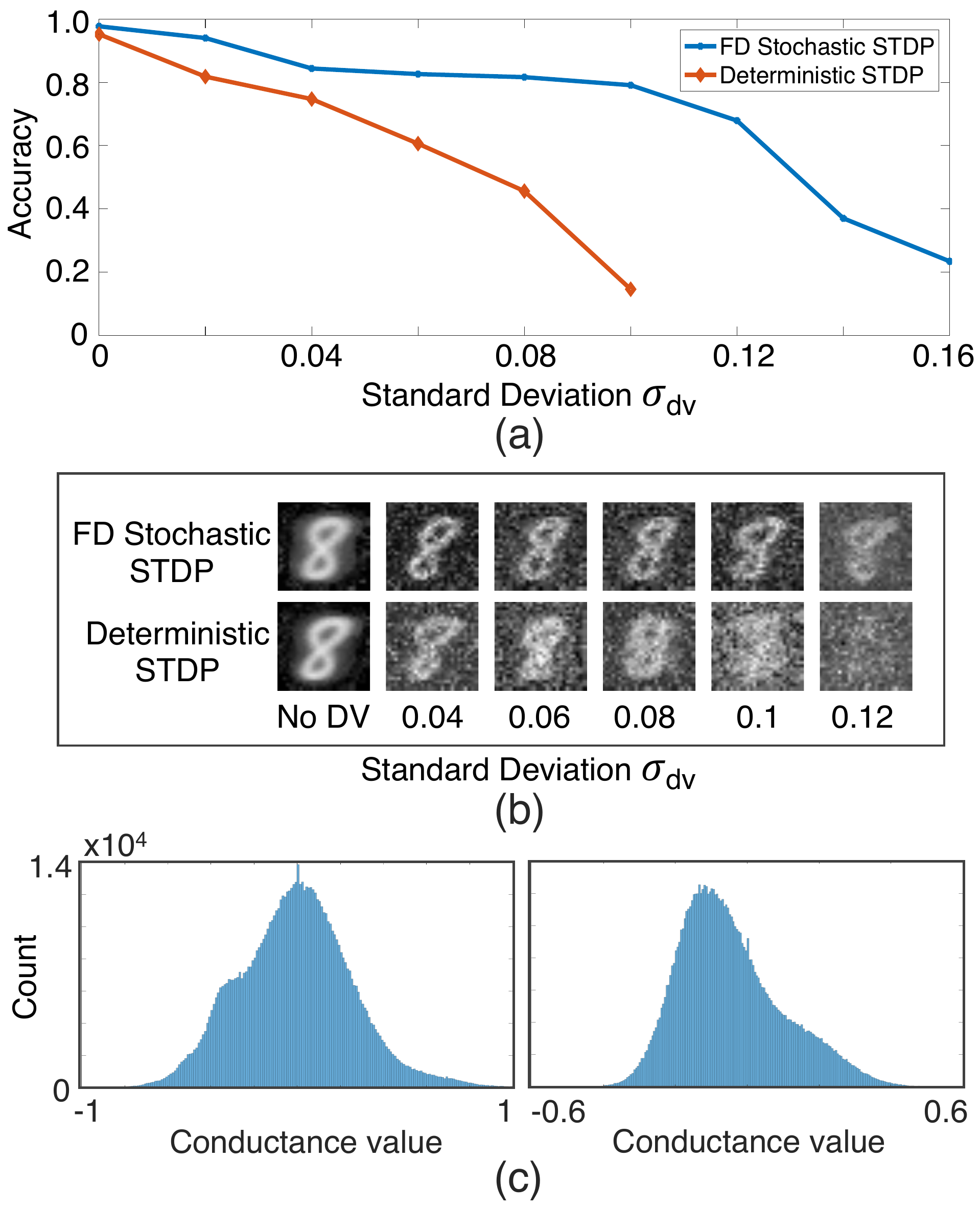}}
    \caption{(a) Accuracy of the network with different standard deviation of device variation. (b) Examples of patterns learned by networks with no device variation (No DV) and different standard deviation of device variation. (c) Distribution of synapse conductance for deterministic STDP (left) and FD stochastic STDP (right) at $\sigma_{dv} = 0.08$.}
\label{fig_dv_1}
\end{figure}

In order to understand quantitatively how ReRAM device variation affects learning ability of SNN, we test networks with a range of variations and compare accuracy of the each setting. In this part, no input images noise is used, and device variation during inference has the same distribution as that for learning. Fig.~\ref{fig_dv_1} (a) shows change in the result accuracy when ReRAM devices are at different levels of variation. Deterministic STDP and FD stochastic STDP both experience accuracy drop. However, accuracy of deterministic STDP drops around 3 times larger than FD stochastic STDP from no variation to a value of $\sigma_{dv} = 0.02$, and degrade faster with increasing device variation. On the other hand, FD stochastic STDP is more robust to device variation. It has the capability to adapt to a wide range of randomness in the resistance modulation process and achieves above 80\% accuracy with up-to $\sigma_{dv} = 0.1$, at which deterministic STDP fails to provide meaningful accuracy. 

FD stochastic STDP provides accuracy improvement as it is able to filter out modulation activities induced from spike pairs that have low causality, or from associative synaptic events, therefore statistically keeps the learned pattern from being disrupted by random modulation values. Apparently, such filtering can maintain a promising level of effectiveness under impact of an extended range of variations until the randomness becomes too high. Fig.~\ref{fig_dv_1} (b) shows example patterns learned by networks with different device variation. Patterns from FD stochastic STDP network experience minor degradation from $\sigma_{dv} = 0.04$ to $\sigma_{dv} = 0.12$, while patterns from deterministic STDP network lose their distinct features rapidly as device variation increases. Fig.~\ref{fig_dv_1} (c) shows conductance distribution of deterministic STDP (left) and FD stochastic STDP based network with device variation of $\sigma_{dv} = 0.08$.

\begin{table}[b]
\centering
\small
\caption{Accuracy results (\%) for learning noisy input (top: deterministic STDP, bottom: FD stochastic STDP) and with device variation from $\sigma_{dv} = 0.04$ to $\sigma_{dv} = 0.14$
.}

\begin{tabular}{c c c c c c c}

      & No DV & 0.04 & 0.08 & 0.1 & 0.12 & 0.14 \\ \hline \hline

no noise&	95.2&	74.7&	45.6	&14.6&	10.3&	10.5\\
10 dB&	89.5&	79.0&	56.0&	19.1&	13.9&	11.7\\
5 dB	&78.3&	78.1&	62.9&	23.9&	17.0&	12.5\\
0 dB&	71.6&	69.6&	60.8&	25.4&	22.7&	14.5\\
\hline

5\%	&92.1&	83.4&	57.3&	19.3&	14.0&	12.3\\
20\%&	76.4&	74.2&	52.0&	22.9&	16.7&	13.2\\
35\%&	68.3&	66.5&	47.6&	15.3&	14.8&	13.9\\
\hline
\end{tabular}
\begin{tabular}{c c c c c c c}
      \\ 
       & No DV & 0.04 & 0.08 & 0.1 & 0.12 & 0.14 \\ \hline \hline

No noise& 97.7&84.4&81.6&79.1&67.9&37.0\\
10 dB&95.9&87.4&83.4&80.4&76.8&43.3\\
5 dB&85.8&85.3&84.0&81.5&78.3&46.7\\
0 dB&78.8&76.8&74.6&76.7&77.2&51.6\\
\hline

5\%&97.2&88.9&85.6&81.1&74.6&42.5\\
20\%&78.1&77.6&76.7&78.1&75.7&49.3\\
35\%&73.2&71.8&71.3&69.8&68.6&49.5\\
\hline
\end{tabular}
\label{table:result_1}
\end{table}

\subsection{Learning with Device Variation and Noisy Input}
In this test, we investigate performance of the proposed algorithm when device variation and input noise are present at the same time. Accuracy is tested with images that have the same type and level of noise as the ones used in learning, and device variation during inference has the same distribution as that for learning. The result is shown in Table~\ref{table:result_1}, with each column showing accuracy under certain device variation $\sigma_{dv}$. FD stochastic STDP outperforms deterministic STDP in all test conditions, and achieves greater accuracy gain at higher device variation. For both algorithm, compared to when no noise is present in the input, accuracy drops slower with increased device variation when input is noisy. Also, comparing results from one specific device variation above zero, maxmium accuracy occurs when input is noisy, instead of when no noise is applied. This behavior shows that when learning noisy input the network experiences increase in the robustness, and smooth out the degradation of accuracy when device variation is present in STDP learning.

\section{Conclusion}
We present FD stochastic STDP as an algorithmic development that serves several purposes. It attempts to addresses the associativity behavior in nervous systems, which has been absent in conventional STDP algorithm used in SNN. It provides a more biologically plausible learning rule while at the same time achieves better performance. FD stochastic STDP also has the benefit of easy adoption and high efficiency as it uses the frequency information readily available in spike-timing based SNN. We show that FD stochastic STDP based SNN can achieve better accuracy when learning noisy input data compared to network based on deterministic STDP. It also exhibits higher robustness to randomness in conductance modulation resulting from device variation in ReRAM based SNN accelerators, thus provides an solution to designing reliable ReRAM based PIM accelerator for SNN algorithmically.

\section*{Acknowledgment}
This work is supported in parts by Office of Naval Research Young Investigator Program, National Science Foundation, and Semiconductor Research Corporation nCORE Program.
\bibliography{reference.bib}{}
\bibliographystyle{unsrt}

\end{document}